\documentclass[runningheads]{llncs}

\usepackage{eccv}

\usepackage{eccvabbrv}

\usepackage{graphicx}
\usepackage{booktabs}

\usepackage[accsupp]{axessibility}  

\usepackage{hyperref}

\usepackage{orcidlink}

\usepackage{float}
\usepackage{xspace}
\usepackage{booktabs}
\usepackage{multirow}
\usepackage{makecell}
\usepackage{amsmath,amssymb,cancel}
\usepackage{enumitem}

\usepackage{graphicx}
\usepackage{caption} 

\usepackage{booktabs,makecell}
\usepackage{multirow}
\usepackage{multicol}
\usepackage{stfloats}
\usepackage{xcolor,colortbl}

\usepackage{xcolor}
\usepackage{comment}
\usepackage[flushleft]{threeparttable}
\usepackage{tcolorbox}
\usepackage{algorithm,algorithmicx,algpseudocode}

\begin{document}

\title{
SnapGen{\small\textsuperscript{++}}: 
Unleashing Diffusion Transformers \\ for Efficient High-Fidelity Image Generation \\ on Edge Devices} 

\titlerunning{SnapGen++}

\author{
Dongting Hu\textsuperscript{1,2}
Aarush Gupta\textsuperscript{1} 
Magzhan Gabidolla\textsuperscript{1} 
Arpit Sahni\textsuperscript{1} \\
Huseyin Coskun\textsuperscript{1} 
Yanyu Li\textsuperscript{1}
Yerlan Idelbayev\textsuperscript{1}
Ahsan Mahmood\textsuperscript{1}\\
Aleksei Lebedev\textsuperscript{1}
Dishani Lahiri\textsuperscript{1}
Anujraaj Goyal\textsuperscript{1} 
Ju Hu\textsuperscript{1}\\
Mingming Gong\textsuperscript{2,3}
Sergey Tulyakov\textsuperscript{1}  Anil Kag\textsuperscript{1}\\
}

\authorrunning{D.~Hu et al.}

\institute{
\textsuperscript{1}~Snap Inc.\quad 
\textsuperscript{2}~University of Melbourne  
\quad \textsuperscript{3}~MBZUAI\\
Project Page: \href{https://snap-research.github.io/snapgenplusplus/}{https://snap-research.github.io/snapgenplusplus/}}

\maketitle

\begin{center}
    \centering
    \captionsetup{type=figure}
    \includegraphics[width=1.0\textwidth]{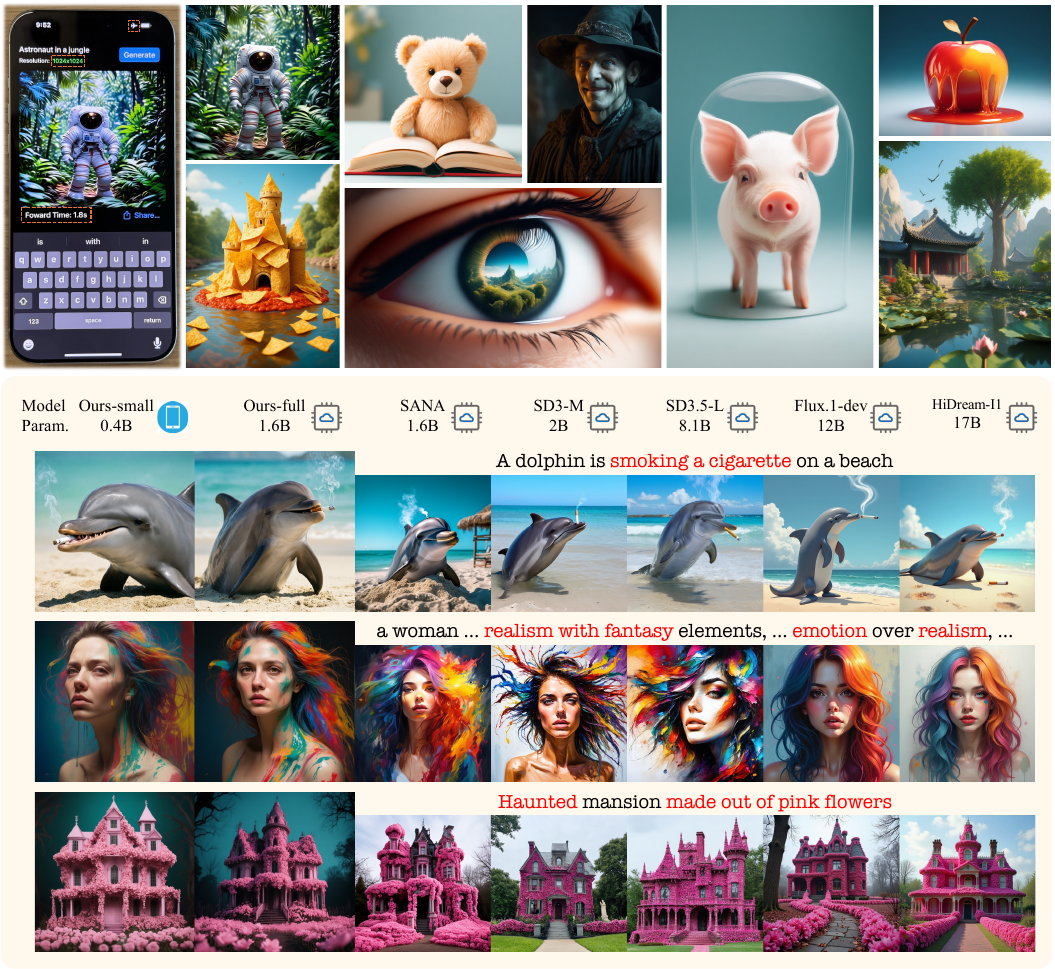}
    \caption{
\textbf{Top:} Our text-to-image Diffusion Transformer (\textit{0.4B} parameters) generates diverse, high-fidelity \textit{1K} images in just \textit{1.8\,s} on a \textit{mobile} device. All examples are produced by this on-device model at a resolution of approximately \textit{1024\textsuperscript{2}}. 
\textbf{Bottom:} Visual comparison across various text-to-image models. Both our on-device (\textit{small}) and server-side (\textit{full}) variants achieve competitive image quality.
}

    \label{fig:teaser}
\end{center}

\begin{abstract}
  Recent advances in Diffusion Transformers (DiTs) have set new standards in image generation, yet their substantial computational and memory demands make them impractical for on-device deployment. In this work, we present an efficient DiT framework tailored for mobile and edge devices, achieving transformer-level generation quality under strict resource constraints. Our approach consists of three key components. First, we design a compact DiT architecture with an adaptive global–local sparse attention mechanism that effectively balances global context modeling and local detail preservation. Second, we introduce an elastic training framework that jointly optimizes sub-DiTs of varying capacities within a unified supernetwork, enabling a single model to dynamically adjust for efficient inference across different hardware. Finally, to produce high-fidelity and low-latency generation (e.g., 4-step) suitable for real-time on-device use, we introduce K-DMD (Knowledge-guided Distribution Matching Distillation), a stability-oriented step-distillation framework that extends the DMD with knowledge transfer from a few-step teacher. Together, these contributions enable efficient, scalable, and high-quality diffusion models for diverse hardware deployment.
\end{abstract}

\section{Introduction}\label{sec:intro}
Image generation models~\cite{rombach2022high,podell2023sdxl,esser2024sd3,blackforestlabs2024flux,wu2025qwenimagetechnicalreport} have made remarkable progress, enabling a broad range of creative applications. Recent advances~\cite{chen2023pixartalpha,peebles2023scalable} show a clear shift toward Diffusion Transformer (DiT) architectures, with large-scale models such as Flux~\cite{blackforestlabs2024flux} and Qwen-Image~\cite{wu2025qwenimagetechnicalreport} achieving state-of-the-art image quality, editing flexibility, and personalization. However, these transformer-based models are extremely large—often containing tens of billions of parameters—and typically require server-grade GPUs and custom CUDA kernels~\cite{li2024svdquant} for inference, resulting in high computational cost and strong reliance on cloud infrastructure. 

To improve accessibility, recent works  such as SnapFusion~\cite{li2024snapfusion}, Mobile Diffusion~\cite{zhao2023mobilediffusion}, and SnapGen~\cite{hu2024snapgen} demonstrate efficient on-device image generation using lightweight U-Net backbones that achieve favorable quality--efficiency trade-offs.
While these on-device models alleviate latency and cloud dependence, their U-Net-based architectures lag far behind recent DiT models in scalability and generative performance. To bridge this architectural gap, we propose an \textbf{Efficient Diffusion Transformer} tailored for mobile and edge deployment, achieving server-level generation quality under strict resource constraints. To address the quadratic complexity of attention especially at high resolutions (e.g., 1K), we introduce a three-stage DiT with an \emph{adaptive global–local sparse attention} mechanism that effectively combines coarse-grained \emph{Key–Value (KV) Compression} for global context modeling with fine-grained \emph{Blockwise Neighborhood Attention} for local spatial relation modeling. By dynamically allocating attention based on content, the model achieves flexible computation and high representational fidelity, outperforming U-Net–based systems such as SnapGen~\cite{hu2024snapgen} in generation quality while maintaining comparable inference speed.

Deploying such models on real-world devices presents another key challenge: hardware heterogeneity. On-device generation must satisfy strict constraints on computation, memory, and power across diverse platforms, from entry-level smartphones to flagship devices and edge servers. A single static model cannot efficiently cover this spectrum, leading to fragmented development and suboptimal deployment. 
To address this, we introduce an \textbf{Elastic Training Framework} for \emph{generative diffusion transformers} that jointly optimizes sub-DiTs of varying capacities within a unified DiT supernetwork. At inference, the appropriate sub-network could be selected dynamically to adapt to different hardware without retraining, enabling scalable and efficient deployment across heterogeneous devices while maintaining consistent generation quality.

To narrow the performance gap between large-scale and compact diffusion models, we adopt a two-stage distillation strategy. We first perform Knowledge Distillation (KD)\cite{hu2024snapgen} from a cloud-scale full-step teacher to transfer generative capability to the compact student model. 
We then apply step distillation for efficient few-step inference. We observe that directly applying the standard Distribution Matching Distillation (DMD) objective~\cite{yin2024onestep,yin2024improved} to compact models often leads to unstable optimization and visual artifacts. To address this issue, we incorporate additional KD supervision from a few-step teacher alongside the DMD objective, which we refer to as \textbf{Knowledge-guided DMD} (K-DMD). This complementary guidance stabilizes training, enabling reliable low-latency, high-fidelity generation for real-time on-device deployment.


In summary, we present a unified framework for scalable on-device image generation with diffusion transformers, integrating architectural efficiency, elastic deployment, and knowledge-guided distillation into a cohesive system. Our approach bridges the long-standing gap between U-Net--based mobile diffusion models and large-scale DiT systems, enabling high-fidelity image generation across heterogeneous hardware platforms. 
Notably, our approach demonstrates that DiTs can run efficiently on edge devices, achieving 1.8s inference on an iPhone 16 Pro Max while generating 1024×1024 images with strong quality (85.2\% on DPG-Bench~\cite{hu2024ella-dpg-bench} and 0.70 on GenEval~\cite{ghosh2024geneval}). 
By jointly addressing model design, deployment flexibility, and training efficacy, our work takes a substantial step toward practical, high-quality on-device generative AI.

\section{Related Work}

\textbf{T2I Diffusion Models.}  
Diffusion models~\cite{ho2020denoising, song2021score, blackforestlabs2024flux, wu2025qwenimagetechnicalreport} have become the state of the art in text-to-image (T2I) generation, surpassing earlier GAN-based approaches~\cite{goodfellow2014generative, brock2018large} in fidelity and diversity. Early latent diffusion models~\cite{rombach2022high, podell2023sdxl, neurips2022imagen, kag2024ascan, li2024snapfusion, chen2024pixart, playground-v1, playground-v2, li2024playground} employed U-Net backbones for iterative denoising in latent space, balancing image quality and memory efficiency. Recent advances replace U-Nets with Diffusion Transformers (DiTs)~\cite{peebles2023scalable, blackforestlabs2024flux, wu2025qwenimagetechnicalreport, wan2025}, achieving improved scalability, quality, and generalization~\cite{liu2023magicedit, wu2025qwenimagetechnicalreport, wan2025,meng2022sdedit, jin2025tpblend}. However, their billion-scale parameters and high computational cost make on-device deployment impractical.

\noindent\textbf{Efficient Diffusion Transformers.}  
Recent efforts~\cite{chen2024pixart, crowson2024hourglass, xie2025sana, liu2024linfusion, li2024svdquant,park2025sprintsparsedenseresidualfusion} aim to improve DiT efficiency.  PixArt-$\Sigma$~\cite{chen2024pixart} introduces key–value compression for 4K image generation, while SANA~\cite{xie2025sana} employs linear self-attention to enable efficient synthesis on consumer GPUs. LinFusion~\cite{liu2024linfusion} replaces quadratic attention in Stable Diffusion~\cite{rombach2022high} with Mamba-based~\cite{mamba, mamba2} attention. Hybrid designs such as Simple Diffusion~\cite{pmlr-v202-hoogeboom23a-simple-diffusion, hoogeboom2025simpler}, HourGlass-DiT~\cite{crowson2024hourglass}, and U-DiTs~\cite{tian2024udits} combine convolutional and transformer blocks in U-Net–style hierarchies. U-ViT~\cite{bao2022all} introduces long-skip connections for faster convergence, and Playgroundv3~\cite{liu2024playground_v3} reduces key/value dimensions to mimic single-level U-Nets. Despite these advances, DiTs still depend on \emph{quadratic attention} and large memory footprints, limiting efficient high-resolution (e.g., 1024$\times$1024) generation on mobile devices.

\noindent\textbf{On-Device Generative Models.}  
To enable on-device deployment, prior works have explored quantization~\cite{BitsFusion, li2024svdquant}, pruning~\cite{li2024snapfusion, hu2024snapgen,wu2025tamingdiffusiontransformerefficient,snapgenv2025yushu}, and knowledge distillation~\cite{hu2024snapgen, kim2023bk} to reduce model size and latency.  
Early on-device systems~\cite{li2024snapfusion, castells2024edgefusion, zhao2023mobilediffusion} pruned and distilled U-Net architectures to generate $512^2$ images within seconds.  
SnapGen~\cite{hu2024snapgen} demonstrated $1024^2$ image generation with a compact U-Net, though with trade-offs in quality and editing flexibility.  
To our knowledge, no prior work has deployed an efficient DiT for high-fidelity on-device generation.

\noindent\textbf{Elastic Networks.}  
Once-for-All~\cite{cai2019onceforall} and Slimmable Networks~\cite{yu2018slimmable} pioneered supernetworks adaptable to varying computational budgets for recognition and detection tasks.  
Follow-up studies~\cite{hanruiwang2020hat, devvrit2024matformer, valipour2024sortednetscalablegeneralizedframework, DynaBert2020} extended this idea to transformers and large language models.  
However, elastic architectures remain underexplored in \emph{generative models}.  
We build our model by introducing an \emph{Elastic DiT} framework that enables flexible diffusion transformer deployment across heterogeneous devices without retraining separate models.

\noindent\textbf{Sparse Attention.}
\cite{yuan2025nsa} and \cite{hassani2025gna} propose hardware-efficient sparse attention designs using block- or neighborhood-based formulations optimized for GPUs.
For video generation, \cite{zhang2025sta} and \cite{xi2025svg} exploit local and spatiotemporal sparsity for efficient attention, while \cite{xia2025adaspa} introduces adaptive sparse attention with online sparsity discovery without retraining. 
Unlike prior these block-wise methods that rely on selective activation or specialised parallel kernels (incompatible with mobile hardware), our method explicitly accounts for both \emph{global} and \emph{local} context modeling, achieving a balanced representation of long-range dependencies and fine-grained details. 
Our local component introduces a unified block-wise \emph{neighborhood} formulation and is tailored for standard on-device deployment. 

\noindent\textbf{Step Distillation.}  
Step distillation accelerates diffusion inference by compressing multi-step sampling into a few denoising iterations~\cite{salimans2022progressive, song2023consistency, lin2024sdxllightning, yin2024improved, xu2024ufogen, bandyopadhyay2025sd35flash}.  
Progressive Distillation~\cite{salimans2022progressive, meng2022on} first showed that student models can learn from intermediate teacher trajectories, while Consistency Models~\cite{song2023consistency, wang2024phased, chen2025sanasprint} enhance stability by enforcing cross-step prediction consistency.  
Adversarial Distillation~\cite{lin2024sdxllightning, xu2024ufogen} introduces GAN-style objectives for few-step, high-fidelity synthesis. Distribution Matching Distillation (DMD)~\cite{yin2024onestep,yin2024improved} aligns teacher–student noise distributions for improved perceptual quality, and has been widely adopted due to its simplicity and effectiveness~\cite{yin2025causvid, huang2025selfforcing, bandyopadhyay2025sd35flash}.   

\section{Method}\label{sec:method}

We introduce an efficient Diffusion Transformer (DiT) architecture, an elastic training framework, and a multi-stage distillation pipeline. Together, these components enable efficient high-fidelity image generation on edge devices.

\vspace{-1em}
\begin{figure*}
    \centering
    \includegraphics[width=\linewidth]{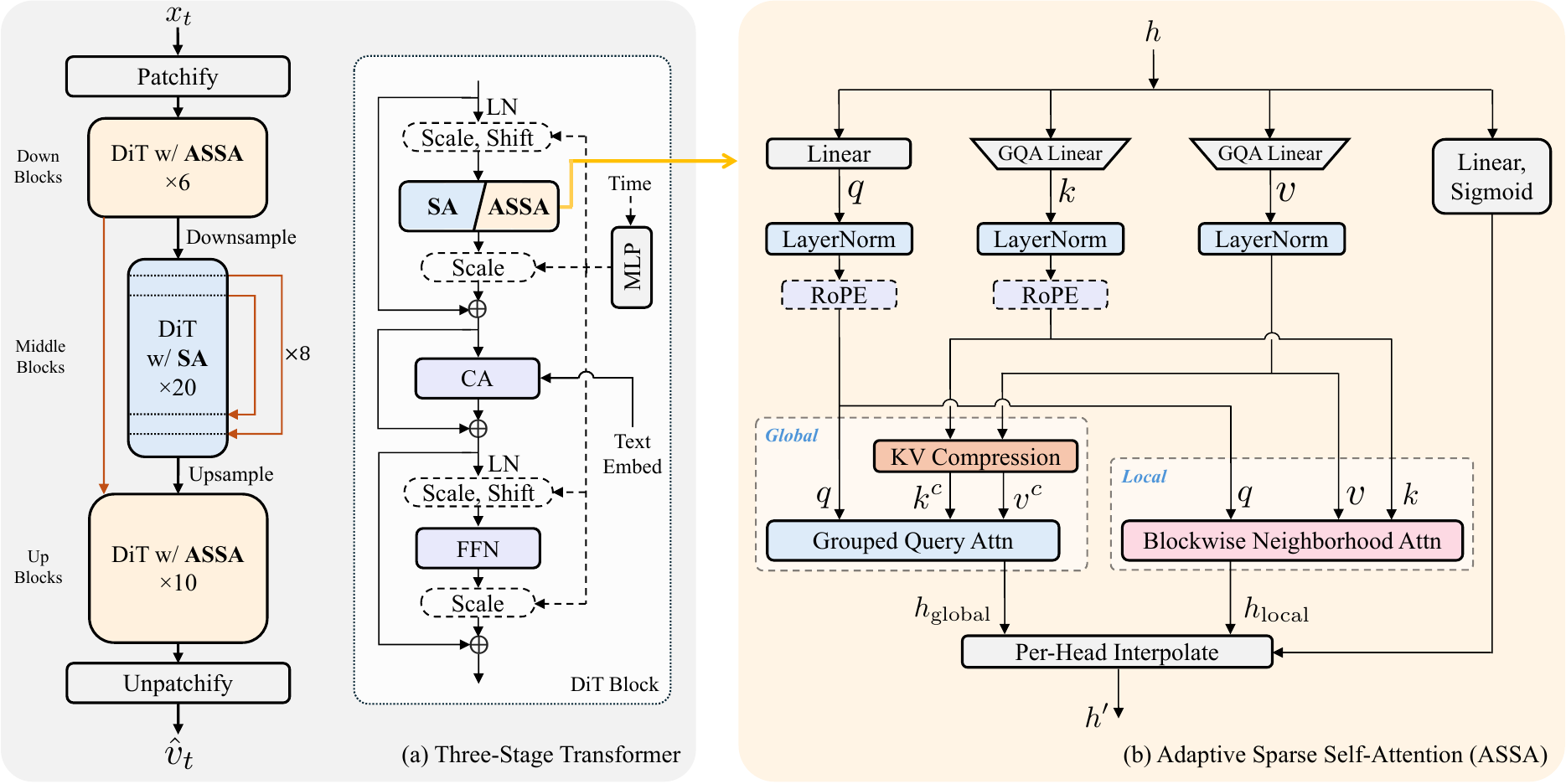}
    \caption{\textbf{Efficient DiT Overview.} \textit{Left}: Our model consists of three stages: \textit{Down, Middle and Up}. Down and Up blocks operate on high-resolution latent while using our novel Adaptive Sparse Self-Attention (ASSA) layers. Middle blocks operate at latents downsampled by $2\times2$ window and use standard Self-Attention (SA) layers. Other layers in the blocks are Cross-Attention (CA) for modulating with input text conditioning and Feed-Forward (FFN) layer. \textit{Right}: Our ASSA layer consists of two parallel attention processing branches: (i) coarse-grained key-value compression for overall structure, and (ii) fine-grained blockwise neighborhood attention features. Finally, the two features are dynamically weighted per head through an input-dependent weighting function.}
    \label{fig:architecture_overview}
    \vspace{-3em}
\end{figure*}

\begin{figure*}
  \centering
    \includegraphics[width=\linewidth]{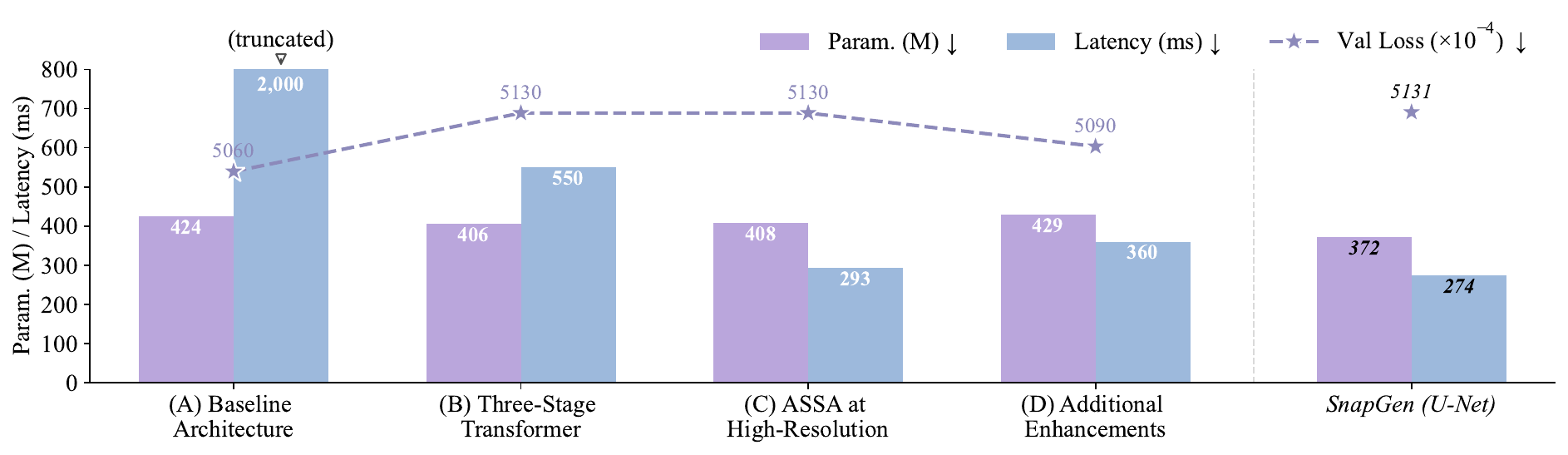}
    \caption{\textbf{Efficient DiT Ablations.}
    We plot the validation loss and model footprint (parameters \& latency on iPhone 16 Pro Max) for various stages in our ablations. Using a baseline DiT yields extremely high latency. Our multi-stage design with ASSA layers and additional enhancements yields an Efficient DiT with comparable latency and better performance than the state-of-the-art on-device model SnapGen~\cite{hu2024snapgen}.}
    \label{fig:dit_ablation}
    \vspace{-1em}
\end{figure*}

\subsection{Efficient DiT Architecture}\label{sec.method_efficient_dit}
Unlike U-Net baselines (e.g., SnapGen~\cite{hu2024snapgen}) that save computation by removing Self-Attention in early stages, DiTs depend on attention throughout the network, necessitating a different strategy. We develop the efficient DiT through a series of key architectural design ablations. All variants are trained on the ImageNet-1K~\cite{deng2009imagenet} for conditional image generation at $256\times256$ resolution and evaluated using the \emph{validation loss} (Val Loss) following the protocol of~\cite{wan2025}. This metric shows stronger correlation with perceptual quality and human preference than conventional metrics like FID~\cite{fid}, particularly under architectural variations, aligned with the findings in~\cite{esser2024sd3}. Model efficiency is measured by parameter count and inference latency on iPhone~16~Pro~Max. For consistency, all models employ the Flux VAE~\cite{blackforestlabs2024flux} and the CLIP-L~\cite{radford2021learning} text encoder, and are trained for 200K iterations using the flow-matching~\cite{liu2022flow, lipman2022flow}. As a reference, we implement the SnapGen~\cite{hu2024snapgen} (\cref{fig:dit_ablation}, rightmost column) as our baseline, which achieves a latency of \textit{274\,ms} and a Val Loss of \textit{0.5131}.

\noindent\textbf{(A) Baseline Architecture.}
Our design builds on the PixArt-$\alpha$~\cite{chen2023pixartalpha} DiT backbone, chosen for its strong balance between parameter efficiency and computational cost. To adapt it for edge deployment, we incorporate multi-query attention (MQA)~\cite{shazeer2019fast-mqa} and reduce the feed-forward network (FFN) expansion ratio to three, yielding a compact 424M-parameter DiT. This baseline attains a validation loss of \textit{0.506} with an inference latency of \textit{2000\,ms} (\cref{fig:dit_ablation}, first column).

\emph{Computation Analysis.}
The main computational bottleneck arises from Self-Attention (SA) at high resolutions. For a $1024^2$ image, the VAE encoder yields a $128^2$ latent map. After patchification, this corresponds to $64^2$ tokens (\textit{4096} in total), substantially increasing SA cost and often causing out-of-memory (OOM) errors on edge hardware. To address this, we introduce below macro- and micro-level modifications to improve efficiency while preserving generation fidelity.

\noindent\textbf{(B) Three-Stage Diffusion Transformer.}
Inspired by recent efficient multi-stage architectures~\cite{crowson2024hourglass,tian2024udits}, we extend the baseline into a three-stage design (\cref{fig:architecture_overview} (a), left). The three stages are denoted as \textit{Down}, \textit{Middle}, and \textit{Up}. A single downsample layer is applied after the down stage and an upsample layer before the up stage, producing a compact latent representation of 1024 tokens ($32\times32$) in the middle stage (bottleneck). Half of the transformer layers are assigned to the middle, while the remaining layers are divided between the down and up—with slightly more layers in the up blocks, inspired by the finding of SiD2~\cite{hoogeboom2025simpler}. This design cuts latency from \textit{2000\,ms} to \textit{550\,ms}, while increasing the validation loss to \textit{0.513} (\cref{fig:dit_ablation}, second column).


\noindent\textbf{(C) Adaptive Sparse Self-Attention (ASSA) at High-Resolution Stages.}
Although the three-stage design reduces the overall computational cost, the primary bottleneck remains in the SA operations in the down and up blocks. To alleviate this, we introduce an Adaptive Sparse Self-Attention (ASSA) (\cref{fig:architecture_overview} (b)) that replaces full SA over \textit{4096} tokens with two complementary components:

\textbf{(i) Global Attention.} We apply Key-Value (KV) compression by performing a $2{\times}2$ convolution with stride 2 on the $k$ and $v$ feature maps. Given the key and value tensors $k, v \in \mathbb{R}^{H\times W\times d}$, we compute the compressed tensors
\begin{equation}
    k^c = \mathrm{Conv}_{2\times2,\,s=2}(k), \quad 
    v^c = \mathrm{Conv}_{2\times2,\,s=2}(v),
\end{equation}
resulting in $k^c, v^c \in \mathbb{R}^{\frac{H}{2}\times \frac{W}{2}\times d}$. 
This reduces the key/value token length by a factor of four, enabling each query to attend to a compressed global context with substantially lower memory and computational overhead.
    
\begin{figure}
    \centering
    \includegraphics[width=0.9\columnwidth]{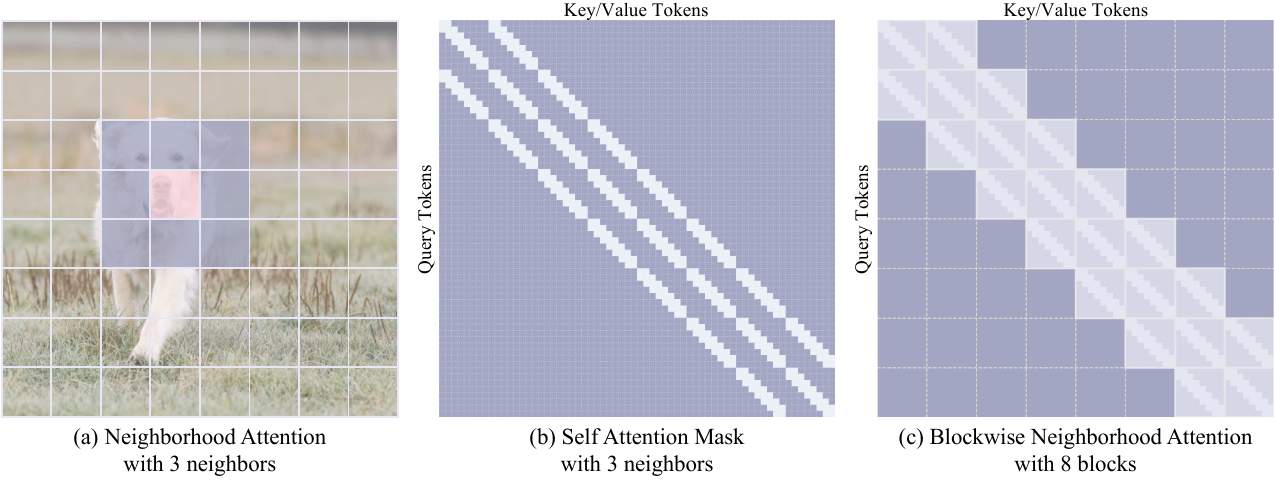} 
    \caption{%
\textbf{Illustration of Blockwise Neighborhood Attention.} 
\textit{(a)} Naive Neighborhood Attention, where each query attends to its local window of 3 neighboring tokens. 
\textit{(b)} Corresponding self-attention mask showing the limited receptive field for each query. 
\textit{(c)} Blockwise Neighborhood Attention extends this concept by grouping tokens into 8 non-overlapping local blocks and attending to neighboring blocks, enabling sparse computation while preserving token-level locality.
}
\vspace{-1em}
    \label{fig:bna_illustration}
\end{figure}

\textbf{(ii) Local Attention.} 
To preserve fine-grained spatial details, we introduce \emph{Blockwise Neighborhood Attention} (BNA), a structured local attention mechanism tailored for mobile-efficient computation.
As shown in \cref{fig:bna_illustration}(a), naive neighborhood attention~\cite{hassani2023neighborhood} restricts each token to attending only to its spatial neighbors within a fixed window (\eg, $3\times3$), analogous to a convolutional receptive field.
When visualized in the attention matrix, this local interaction pattern forms a band-diagonal structure, as shown in \cref{fig:bna_illustration}(b). 
While such localized attention is more efficient than full self-attention, it is \emph{not} natively supported on mobile hardware and still incurs nontrivial overhead due to its fine-grained per-token operations. 
To better align with mobile hardware, we first adopt a blockwise formulation, where the token grid is divided into $B$ (a hyperparameter) non-overlapping spatial blocks. 
Specifically, the query, key, and value matrices
$q, k, v \in \mathbb{R}^{(H\!W)\times d}$ 
are partitioned along the sequence dimension into $B$ blocks:
\begin{equation}
q = [q_1; \dots; q_B], k = [k_1; \dots; k_B], v = [v_1; \dots; v_B],
\end{equation}
where each block $q_b, k_b, v_b \in \mathbb{R}^{N_b \times d}$ 
and block size $N_b = HW / B$. 
Rather than restricting attention strictly within each block or dynamically activating certain blocks~\cite{xia2025adaspa, yuan2025nsa}, we introduce a \emph{unified neighborhood formulation}: for each query block $q_b$, 
attention is computed over a neighboring set of  key--value blocks 
$\mathcal{N}_r(b) = \{\,b - r, \dots, b, \dots, b + r\,\}$, 
where $r$ denotes the block neighborhood radius (bandwidth). 
The resulting Blockwise Neighborhood Attention is defined as
\begin{equation}
A_b = 
\operatorname{Softmax}\!\left(
    \frac{q_b [k_{\mathcal{N}_r(b)}]^{\top}}{\sqrt{d}}
\right)
[v_{\mathcal{N}_r(b)}],
\quad b = 1, \dots, B,
\end{equation}
where $[k_{\mathcal{N}_r(b)}]$ and $[v_{\mathcal{N}_r(b)}]$ 
represent the concatenation of key and value blocks within the neighborhood $\mathcal{N}_r(b)$. 
As illustrated in \cref{fig:bna_illustration}(c), this formulation recovers token-level locality and can extend the effective receptive field beyond a fixed window, with complexity $\mathcal{O}(N^2 / B)$.
Compared to per-token neighborhood attention, which incurs irregular memory access and high overhead on mobile devices, the blockwise neighborhood formulation leverages contiguous token chunks for efficient batched computations and improved hardware utilization. 
Different choices of block number $B$ and neighborhood radius $r$ allow flexible control of the token-level receptive field (see supplementary material for details).

\medskip
\noindent
The final attention score is a linear interpolation between global attention and local attention, conditional on the input hidden states. This novel sparse attention design substantially reduces the overall attention overhead while preserving generation quality. 
As shown in \cref{fig:dit_ablation} (third column), our sparse attention model achieves a latency of \textit{293\,ms} without loss of generation quality (val loss of \textit{0.513}).\clearpage

\noindent\textbf{(D) Additional Enhancements.}
We further apply the following refinements:
\begin{itemize}
    \item \textbf{Dense long-range skip connections:} Following~\cite{bao2022uvit}, we add dense skip connections in the middle stage to increase the capacity of the bottleneck.
    
    \item \textbf{Grouped Query Attention (GQA):} We employ GQA~\cite{ainslie-etal-2023-gqa} by increasing the number of key/value heads to eight, improving multi-head diversity and reducing query–key bottlenecks with minimal parameter overhead.
    
    \item \textbf{Expanded FFN capacity:} The FFN expansion ratio increases to four in down and up stages, improving representational capacity without incurring significant additional computational cost.
    
    \item \textbf{Layer redistribution:} Four transformer layers are reassigned from the middle stage—two each to the down and up—to achieve a more balanced depth and better information hierarchy. Thanks to the efficiency of the proposed ASSA, this adjustment incurs only a modest computational overhead.

\end{itemize}

As shown in \cref{fig:dit_ablation} (fourth column), this configuration achieves a latency of \textit{360\,ms} and a validation loss of \textit{0.509}, offering a strong trade-off between efficiency and accuracy.
With all components combined, our efficient DiT architecture attains conv-level latency while surpassing it in both visual quality and scalability in image generation, outperforming SnapGen~\cite{hu2024snapgen} by a large margin in validation loss. Some qualitative results are in the supplementary material.




\subsection{Elastic DiT Framework}\label{sec.method_elastic_framework}
Recent works such as Matformer~\cite{devvrit2024matformer} and Gemma-3n~\cite{gemma_3n_2025} demonstrate the importance of building unified yet adaptable LLM architectures that can be deployed efficiently across heterogeneous platforms (e.g., high-end smartphones, low-power devices, and server-side environments). 
Motivated by this, we design an \textit{Elastic DiT} framework that enables a single diffusion transformer to flexibly scale its capacity according to available computational resources.

\noindent\textbf{Framework Design.}
To enable such flexibility, we identify a structural decomposition that supports parameter sharing across subnetworks of varying widths~\cite{yu2018slimmable}, slicing the projection matrices in the attention and FFN layers along the \textit{hidden dimension} to sample subnetworks of varying sizes from a single supernetwork.
In Cross-Attention layers, the key and value projections remain unsliced, as they are independent of the model width. 
Parameters strictly tied to the hidden-state length—such as those in \textit{layer normalization} and \textit{modulation layers}—are kept separate, due to their lightweight and dimension-specific nature. 
This design produces three model variants: a \textit{tiny} 0.3B  model (0.375$\times$ width) for low-end Android devices, a \textit{small} 0.4B model (0.5$\times$ width) for high-end smartphones, and a \textit{full} 1.6B supernetwork (1$\times$ width) that can be quantized for on-device deployment or server-side inference. \\

\begin{figure}[t]
    \centering
    \begin{minipage}{0.47\columnwidth}
        \centering
        \includegraphics[width=\linewidth]{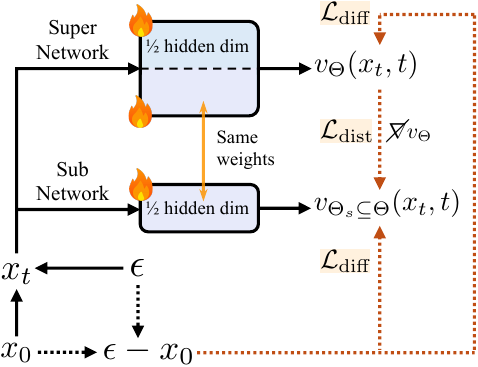}
        \caption{\textbf{Elastic Training Framework.} 
        Given a supernetwork, we define sub-networks as different granularities of the hidden dimension. 
        During training, we sample sub-networks uniformly and supervise them using the output from the supernetwork. 
        In addition, we use standard diffusion loss on all granularities.}
        \label{fig:elastic_training}
    \end{minipage}
    \hfill
    \begin{minipage}{0.47\columnwidth}
        \centering
        \includegraphics[width=\linewidth]{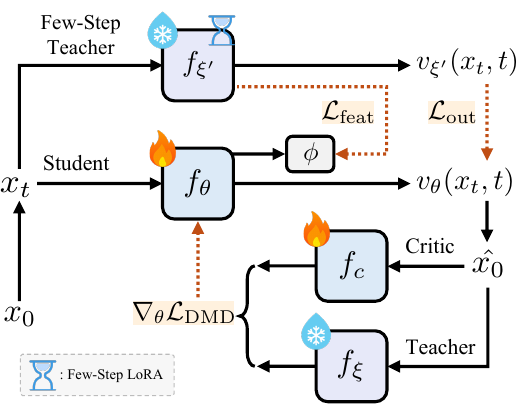}
        \caption{\textbf{Knowledge-guided Distribution Matching Distillation (K-DMD).} 
        Our step distillation method combines the distribution matching objective from a full-step teacher with knowledge transfer from a few-step teacher (implemented via LoRA).}
        \label{fig:kdmd}
    \end{minipage}
    \vspace{-1em}
\end{figure}


\noindent\textbf{Training Recipe.}
Naively optimizing multiple subnetworks with shared weights often leads to unstable gradient updates, even under low learning rates. 
To mitigate this issue, we propose a unified elastic training strategy that stabilizes joint optimization across subnetworks of different widths (\cref{fig:elastic_training}). 
At each iteration, subnetworks parameterized by $\Theta_s \subseteq \Theta$ are sampled alongside the full supernetwork $\Theta$ and optimized under a shared flow-matching objective:
{
\begin{equation}\label{eq:loss_diff}
\mathcal{L}_{\mathrm{diff}}(\theta)
= \mathbb{E}_{\epsilon \sim \mathcal{N}(0,I),\, t}
\Big[ \| (\epsilon - x_0) - v_{\theta}(x_t, t) \|_2^2 \Big],
\end{equation}
}
where $\theta \in \{\Theta,\, \Theta_s\}$.
Their gradients are then aggregated using adaptive scaling to ensure balanced updates across subnetworks.
Additionally, a lightweight distillation loss is applied between each subnetwork and the supernetwork to further improve training stability and ensure consistent convergence behavior:
{
\begin{equation}\label{eq:loss_dist}
\mathcal{L}_{\mathrm{dist}}(\Theta_s)
= \big\| v_{\Theta_s}(x_t, t) - \cancel{\nabla}v_{\Theta}(x_t, t) \big\|_2^2,
\end{equation}
}
where $\cancel{\nabla}$ denotes the stop-gradient operator.
This elastic training framework enables DiT models to be deployed seamlessly across heterogeneous platforms while maintaining strong performance and visual fidelity. 
As shown in \cref{tab:elastic_table}, the elastic training recipe achieves comparable validation loss and DINO-FID to standalone training while reducing the overall model-state footprint through parameter sharing. 
Note that these results are obtained from relatively small-scale experiments on ImageNet, where the overhead from data loading and embedding computation is limited. 
When scaling to large-scale text-to-image (T2I) training and distillation, this overhead becomes significantly more pronounced, as the data pipeline and larger teacher components dominate the total training cost.

\begin{table}[!htbp] 
\vspace{-1em}
\centering
\setlength{\tabcolsep}{4pt}
\caption{Comparison between \textbf{Standalone} and \textbf{Elastic} training for 0.4B and 1.6B models. 
Elastic training reuses parameters between model scales, reducing memory allocation while maintaining similar validation loss and DINO-FID. }
\resizebox{0.9\linewidth}{!}{
\begin{tabular}{lcccc}
\toprule
\textbf{Training Recipe} &
\textbf{Model} &
\textbf{Val Loss} &
\textbf{DINO FID}&
\textbf{Training Footprint} \\
\midrule
\multirow{2}{*}{Standalone}
& 0.4B & 0.5090 & 128 & 6.6 GB \\
& 1.6B   & 0.5073 & 109 & 18.8 GB \\
\midrule
\multirow{2}{*}{Elastic}
& 0.4B & 0.5093 & 125 & -- \\
& 1.6B   & 0.5071 & 110 & 18.8 GB \\
\bottomrule
\end{tabular}
}
\label{tab:elastic_table}
\vspace{-2em}
\end{table}

\subsection{Distillation Pipelines}\label{sec.method_distillation}
During the pretraining stage, we apply both the flow-matching loss~(\cref{eq:loss_diff}) and the elastic distillation loss~(\cref{eq:loss_dist}). 
Following the SnapGen~\cite{hu2024snapgen} pipeline, we then adopt a two-stage distillation strategy:
(i) large-scale knowledge distillation (KD) from a full-step teacher to enhance the performance of compact student models; 
(ii) step distillation to reduce sampling steps, enabling efficient inference and real-time generation on edge devices. \\

\noindent\textbf{Knowledge Distillation.} 
A large cloud-scale teacher~\cite{wu2025qwenimagetechnicalreport} (denoted as $\xi$) supervises the training of the elastic DiT models through both output- and feature-level distillation. 
The student $\theta \in \{\Theta,\, \Theta_s\}$ is first encouraged to match the teacher’s velocity predictions:
{
\begin{equation}\label{eq:loss_output}
\mathcal{L}_{\mathrm{out}}^{\xi}(\theta)
= \big\| v_{\xi}(x_t, t) - v_{\theta}(x_t, t) \big\|_2^2,
\end{equation}
}
and further aligns its internal representations via feature distillation:
{
\begin{equation}\label{eq:loss_feat}
\mathcal{L}_{\mathrm{feat}}^{\xi}(\theta, \phi)
= \big\| f_{\xi}(x_t, t) - \phi \big(f_{\theta}(x_t, t)\big) \big\|_2^2,
\end{equation}
}
where $\phi$ is the projector on the final transformer layer.
The overall distillation objective combines both levels of supervision with timestep-aware scaling~\cite{hu2024snapgen}:
{
\begin{equation}\label{eq:loss_kd}
\mathcal{L}_{\mathrm{KD}}(\theta, \phi)
= \mathcal{S} \big(\mathcal{L}_{\mathrm{diff}},\, \mathcal{L}_{\mathrm{out}}^{\xi}\big)
+ \mathcal{L}_{\mathrm{feat}}^{\xi},
\end{equation}
}
where $\mathcal{S}(\cdot)$ the timestep-aware scaling operator.\\


\noindent\textbf{Step Distillation.}
Following recent few-step distillation methods~\cite{yin2024onestep, yin2024improved}, we adopt Distribution Matching Distillation (DMD) for step distillation, where the KL divergence is computed between the real score from the full-step teacher 
$\xi$ and the student distribution (fake score)
estimated by a critic model $c$:
\begin{equation}\label{eq:loss_dmd}
\begin{aligned}
\nabla_{\theta} \mathcal{L}_{\mathrm{DMD}}^{\xi}(\theta)
&=
\Big[f_c\big(\mathcal{F}(\hat{x}_0,\tau),\tau\big)-f_\xi\big(\mathcal{F}(\hat{x}_0,\tau),\tau\big)\Big]
\frac{d\hat{x}_0}{d\theta},\\[-1pt]
\text{with}\quad \hat{x}_0 &= x_t - \sigma_t\, v_\theta(x_t,t),
\end{aligned}
\end{equation}
where $\tau$ is randomly sampled to diffuse (via $\mathcal{F}$) the input $\hat{x}_0$ before passing into teacher $\xi$ and critic $c$.  
However, standard DMD is sensitive to hyperparameter choices such as teacher guidance scale and auxiliary loss weight. We observe that optimal settings vary across model capacities, and applying to compact models with only millions of parameters often causes unstable convergence (\cref{fig:dmdvskdmd}).

To address this issue and leverage rich knowledge from cloud-scale teachers, we introduce Knowledge-guided DMD (K-DMD), a practical stability-oriented extension of DMD that incorporates additional knowledge transfer from a few-step teacher~\cite{qwen_image_lightning_2025} (denoted as $\xi^\prime$) (\cref{fig:kdmd}).
Specifically, we feed the same input $x_t$ as the student and incorporate $\mathcal{L}_{\mathrm{out}}^{\xi^\prime}$ 
and $\mathcal{L}_{\mathrm{feat}}^{\xi^\prime}$ (\cref{eq:loss_output,eq:loss_feat}) into the 
training objective:
\begin{equation}\label{eq:loss_kdmd}
\mathcal{L}_{\mathrm{K\text{-}DMD}}(\theta, \phi) 
= \mathcal{L}_{\mathrm{DMD}}^{\xi} 
+ \mathcal{L}_{\mathrm{out}}^{\xi^\prime}
+ \mathcal{L}_{\mathrm{feat}}^{\xi^\prime}.
\end{equation}
This combined objective stabilizes training across models of varying capacities without requiring additional hyperparameter tuning, which is particularly important for our compact elastic networks. Rather than introducing a new distillation paradigm, our goal is to integrate complementary supervision signals that support scalable and reliable deployment under diverse hardware constraints.

In practice, the few-step teacher is activated by enabling the few-step LoRA~\cite{hu2022lora} with no additional memory overhead (\cref{fig:kdmd}). The critic $c$ is updated with flow-matching (\cref{eq:loss_diff}) on $\hat{x_0}$ aligned with previous works~\cite{yin2025causvid, huang2025selfforcing, bandyopadhyay2025sd35flash}.
We note that similar supervision strategies could be incorporated into other step distillation methods; here we focus on DMD due to its simplicity and widespread adoption.





\section{Experiments}\label{sec:experiments}
In this section, we evaluate our framework on generation quality (\cref{tab:quantitative_benchmarks}, \cref{fig:human_eval}), deployment scalability (\cref{tab:diverse_devices}), and step distillation effectiveness (\cref{fig:dmdvskdmd,fig:fewstep}).
\subsection{Experimental Setup}
\textbf{T2I Configuration.} We use the 1.6B parameter efficient DiT (\cref{sec.method_efficient_dit}) as the supernetwork for our elastic training (\cref{sec.method_elastic_framework}) which embeds two sub-networks of $0.3$B and $0.4$B parameters. We employ TinyCLIP~\cite{tinyclip} and Gemma3-4b-it~\cite{gemmateam2025gemma3technicalreport} as text encoders with token-wise concatenation for rich semantic embeddings. 
Since we use Qwen-Image~\cite{wu2025qwenimagetechnicalreport} as our teacher, we use their VAE to align the latent space. We also train a tiny decoder similar to \cite{hu2024snapgen} for on-device generation.

\noindent\textbf{Training Recipe.} Inspired by recent works~\cite{wu2025qwenimagetechnicalreport}, we use multi-aspect ratio data to pre-train the elastic model using flow-matching loss~\cite{liu2022flow,esser2024sd3} at $256$ resolution, followed by $1024$ base resolution. In the next stage, we use knowledge distillation from Qwen-Image~\cite{wu2025qwenimagetechnicalreport} and K-DMD step-distillation training with Qwen-Image-Lightning~\cite{qwen_image_lightning_2025}. We provide more details in the supplementary material.

\noindent\textbf{On-Device Runtime.} The VAE decoder takes $120\,ms$, while the DiT (0.4B) has a per-step latency of $360\,ms$, resulting in a total runtime of about $1.8\,s$ for 4-step generation on the iPhone 16 Pro Max, including system overhead. Further details of all variants are provided in the supplementary material.


\subsection{Evaluations}
\textbf{Quantitative Results.} We evaluate our full-step models against standard baselines on DPG-Bench~\cite{hu2024ella-dpg-bench}, GenEval~\cite{ghosh2024geneval}, and T2I-CompBench~\cite{huang2025t2icompbench++} to assess key generation attributes. Following~\cite{hu2024snapgen}, we also report CLIP-Score~\cite{radford2021learning} on a subset of MS-COCO~\cite{lin2014microsoft}. Results for the tiny ($0.3$B), small ($0.4$B), and full ($1.6$B) variants of our elastic model, together with several representative prior models, are shown in \cref{tab:quantitative_benchmarks}, with main findings summarized below:

\begin{itemize}
    \item Across all benchmarks, our models demonstrate strong performance at multiple scales. The full 1.6B variant achieves results comparable to large server-side models such as Flux.1-dev~\cite{blackforestlabs2024flux} and SD3.5-Large~\cite{sd3.5}, while the on-device 0.4B small variant remains highly competitive despite notably reduced size.
    \item The small variant (0.4B) surpasses models up to 8B parameters (20$\times$) on multiple benchmarks while retaining on-device efficiency comparable to SnapGen, and the tiny variant (0.3B) achieves the highest throughput.
    \item The elastic design enables a controllable quality–efficiency trade-off within a single unified model, allowing different width configurations to adapt to diverse hardware constraints without retraining.
\end{itemize}

\begin{table}[ht]
\caption{\textbf{Quantitative Evaluation.} 
Scores are reported on DPG-Bench, GenEval, T2I-CompBench, and CLIP (COCO). The number of model parameters is reported in billions.
We use the default inference step count (NFE) for each model. Throughput/FPS (samples/s) is measured on a single 80GB A100 GPU using the largest batch size that fits for $1024\times1024$ images, while latency (ms) is measured on an iPhone 16 Pro Max with one single forward pass.
Best, second-best, and third-best results are marked by \textbf{bold}, \underline{underline}, and \textit{italics}, respectively.}
\label{tab:quantitative_benchmarks}
\centering
\resizebox{\linewidth}{!}{
\begin{tabular}{l|ccc|cc|cccc}
\toprule
Model & Arch. & Param. & NFE &
FPS $\uparrow$ &
Latency$\downarrow$ &
DPG $\uparrow$ & GenEval $\uparrow$ &
T2I-C.B. $\uparrow$ &
CLIP $\uparrow$ \\ 
\midrule
SnapGen~\cite{hu2024snapgen}       & U-Net   & 0.4 & 28 & 0.51 & 274 & 81.1 & 0.66 & -- & \textit{0.332} \\
PixArt-$\alpha$~\cite{chen2023pixartalpha}   & DiT     & 0.6 & 20 & 0.42 & $\dagger$ & 71.1 & 0.48 & 0.351 & 0.316 \\
PixArt-$\Sigma$~\cite{chen2024pixart}        & DiT     & 0.6 & 20 & 0.46 & $\dagger$ & 80.5 & 0.53 & 0.427 & 0.317 \\
SANA~\cite{xie2025sana}                      & Hybrid  & 1.6 & 20 & 0.91 & $\dagger$ & 84.8 & 0.66 & 0.476 & 0.327 \\
LUMINA-Next~\cite{zhuo2024lumina}            & DiT     & 2.0 & 30 & 0.06 & $\dagger$ & 74.6 & 0.46 & 0.353 & 0.309 \\
SD3-Medium~\cite{esser2024sd3}               & DiT     & 2.0 & 28 & 0.28 & $\dagger$ & 84.1 & 0.68 & 0.522 & 0.323 \\
SDXL~\c ite{podell2023sdxl}                   & U-Net   & 2.6 & 50 & 0.18 & $\dagger$ & 74.7 & 0.55 & 0.402 & 0.301 \\
Playgroundv2.5~\cite{li2024playground}       & DiT     & 2.6 & 50 & 0.18 & $\dagger$ & 75.5 & 0.56 & 0.237 & 0.319 \\
IF-XL~\cite{deepfloyd}                       & U-Net   & 5.5 & 50 & 0.06 & $\dagger$ & 75.6 & 0.61 & 0.421 & 0.311 \\
SD3.5-Large~\cite{sd3.5}                     & DiT     & 8.1 & 28 & 0.08 & $\dagger$ & \underline{85.6} & \underline{0.71} & \underline{0.507} & \textit{0.326} \\
Flux.1-dev~\cite{blackforestlabs2024flux}      & DiT     & 12  & 50 & 0.04 & $\dagger$ & 83.8 & 0.66 & 0.471 & 0.316 \\
\midrule
Ours-tiny  & DiT & 0.3 & 28 & 0.81 & 280 & 84.6 & 0.69 & 0.502 & 0.330 \\
Ours-small & DiT & 0.4 & 28 & 0.62 & 360 & \textit{85.2} & \textit{0.70} & \textit{0.506} & \underline{0.332} \\
Ours-full  & DiT & 1.6 & 28 & 0.28 & 1580$^*$ & \textbf{87.2} & \textbf{0.76} & \textbf{0.536} & \textbf{0.338} \\
\bottomrule
\end{tabular}}
{\footnotesize \textit{Note.} “$\dagger$”: out-of-memory (OOM) at $1024^2$ resolution. “$*$”: 4-bit weight quantized}
\end{table}

\noindent\textbf{Qualitative Results.} 
To visually assess image–text alignment and overall aesthetics, we compare images generated by different models in \cref{fig:teaser}. Despite their compact size, our models—benefiting from knowledge distillation—remain robust in following prompts and capturing complex concepts. In contrast, many existing models often produce overly stylized or less realistic images, failing to fully reflect the prompt and occasionally omitting important visual elements. We include more comparisons in the supplementary material. \\

\begin{figure}[h]
\centering
\includegraphics[width=\textwidth]{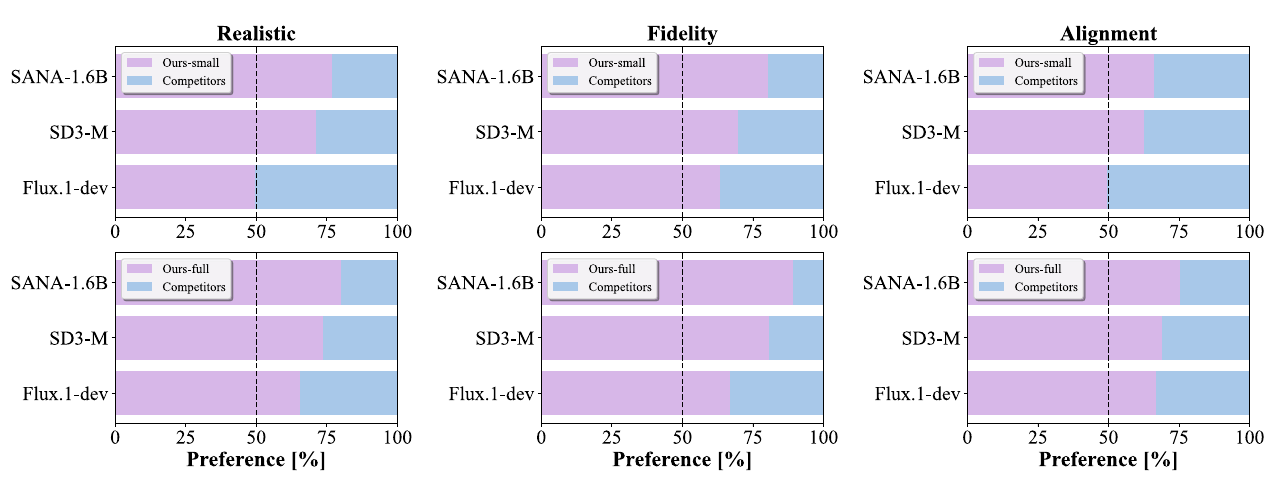}
\vspace{-1em}
\caption{\textbf{Human Evaluation.} A user study comparing our small (0.4B) and full (1.6B) variants with three baselines—SANA (1.6B), SD3-Medium (2B), and Flux.1-dev (12B)—across three key attributes: realism, visual fidelity, and text–image alignment.}
\label{fig:human_eval}
\end{figure}

\noindent\textbf{Human Preference Study.}
For a thorough comparison between baselines, we conduct a user study following the widely used Parti prompts~\cite{yu2022scaling}. We include SANA (1.6B), SD3-M (2B), and Flux.1-dev (12B) as the baselines and ask participants to select images with better attributes between the baselines and our models. The evaluation considers three key aspects: realism, visual fidelity, and text alignment. 
As shown in \cref{fig:human_eval}, our full variant surpasses all baselines in fidelity, realism and image–text alignment. The small variant also demonstrates robust performance, outperforming baselines such as SANA and SD3-M on most attributes while remaining competitive with Flux.1-dev.\\






\noindent\textbf{Cross-Device Deployment.}
To validate the hardware adaptivity enabled by our Elastic Training Framework, we measure the inference latency of sub-networks (small and tiny variants) across multiple commercial mobile devices (\cref{tab:diverse_devices}).

\begin{table}
\centering
\setlength{\tabcolsep}{4pt}
\caption{Forward latency (ms) of our small and tiny variants on different devices.}
\label{tab:diverse_devices}
\resizebox{\linewidth}{!}{
\begin{tabular}{lccccc}
\toprule
Model & Param. & iPhone 16PM & iPhone 14PM & Samsung S25 & Samsung S23 \\
\midrule
Ours-tiny & 0.3B & 280  & 413  & 751  & 998  \\
Ours-small & 0.4B & 360  & 531  & 1125 & 1443 \\
\bottomrule
\end{tabular}
}
\end{table}

\begin{figure}
    \centering
    \begin{minipage}[t]{0.49\linewidth}
        \centering
        \includegraphics[width=\linewidth]{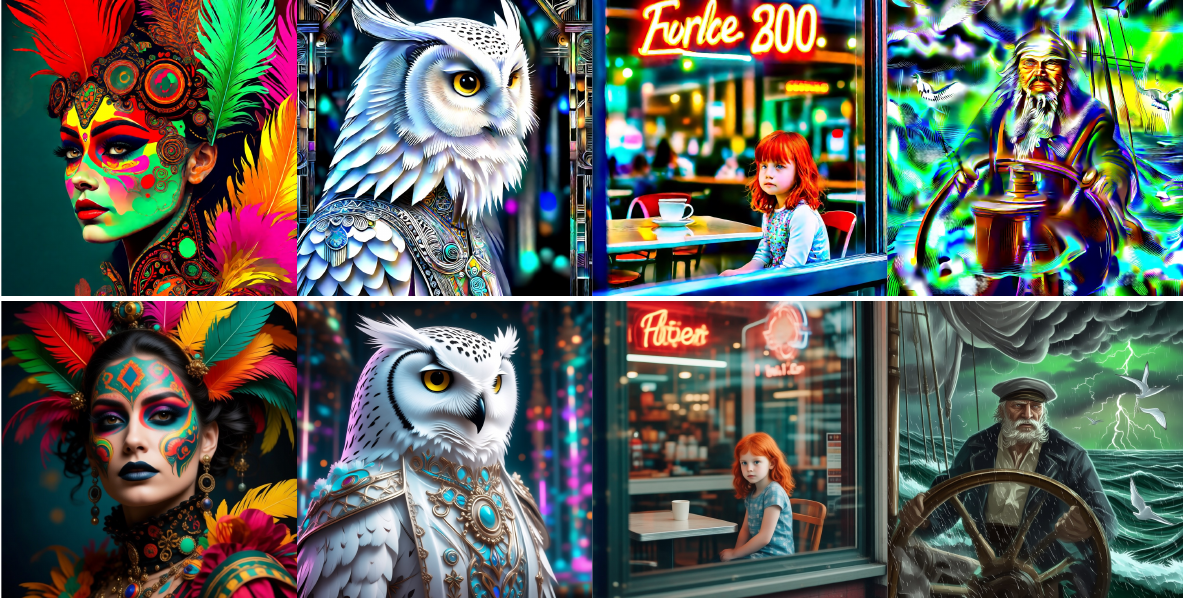}
        \caption{\textbf{DMD vs.\ K-DMD.} By introducing KD regularization, K-DMD (bottom) mitigates the over-saturation and grainy artifacts observed with standard DMD (top) on compact models.}
        \label{fig:dmdvskdmd}
    \end{minipage}
    \hfill
    \begin{minipage}[t]{0.48\linewidth}
        \centering
        \includegraphics[width=\linewidth]{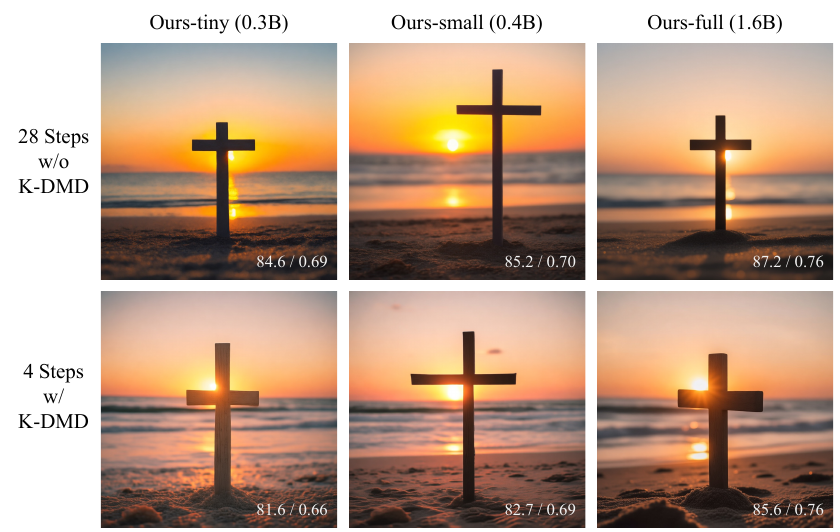}
        \caption{\textbf{Few-step Generation.} 
        Comparison of tiny (0.3B), small (0.4B), and full (1.6B) models under 28-step (w/o K-DMD) and 4-step (w/ K-DMD) settings. Numbers denote DPG / GenEval scores.}
        \label{fig:fewstep}
    \end{minipage}
\end{figure}

\noindent\textbf{Few-Step Generation.}
After applying Knowledge-guided Distribution Matching Distillation (K-DMD), our models are capable of generating high-quality images in only four steps.
As shown in \cref{fig:fewstep}, we compare the performance of the 28-step base models with the 4-step distilled models using DPG-Bench and GenEval scores.
The results indicate that the distilled 4-step models achieve performance comparable to the 28-step baselines, despite the significant reduction in sampling steps.
Importantly, by introducing KD-based regularization, K-DMD mitigates the over-saturation and grainy artifacts commonly observed when applying standard DMD to compact models (\cref{fig:dmdvskdmd}). Overall, although a slight drop in scores is observed, the quality remains largely preserved, highlighting the effectiveness of our step-distillation approach.


\section{Conclusion}\label{sec:conclusion}
In this work, we present an efficient Diffusion Transformer framework that enables high-quality image generation on mobile and edge devices. 
Our architecture combines a multi-stage design, adaptive global–local sparse attention, and several lightweight optimizations to reduce computational and memory overhead while preserving fidelity under strict resource constraints. 
To support scalable deployment, we propose an elastic training framework that unifies width-adjustable subnetworks within a single supernetwork, allowing dynamic adaptation to heterogeneous hardware. 
Furthermore, we introduce Knowledge-guided Distribution Matching Distillation (K-DMD), which stabilizes step distillation in compact models and enables reliable few-step generation for low-latency inference.
Extensive experiments demonstrate that our models achieve competitive quality compared to large server-side systems while operating efficiently on-device. 
Together, these contributions establish a comprehensive and scalable approach for bringing diffusion transformers to real-world edge deployment.



%
%
\bibliographystyle{splncs04}
\bibliography{main}

\clearpage
\author{}
\institute{}
\title{Supplementary Material for SnapGen{\small\textsuperscript{++}}: \\
Unleashing Diffusion Transformers for Efficient High-Fidelity Image Generation on Edge Devices}
\titlerunning{ }
\maketitle

\setcounter{section}{0}
\setcounter{figure}{0}
\setcounter{table}{0}
\renewcommand{\thesection}{\Alph{section}}
\setcounter{section}{0}


\section{Discussion of On-Device Latency}
We report the per-step denoising latency and total generation time in \cref{tab:supp_latency}.
Note that the VAE decoder requires approximately $120\,ms$~\cite{hu2024snapgen}, and additional components such as latent scaling, scheduler stepping, and CLIP embedding introduce negligible latency, similar to observations in~\cite{li2024snapfusion, hu2024snapgen}. Thanks to our proposed Adaptive Sparse Self-Attention, the quantized full model runs on mobile devices without out-of-memory issues, even with attention computed over 4096 tokens at a dimension of 1024.

\begin{table}[!htbp]
  \centering
  \setlength{\tabcolsep}{6pt}
  \caption{Latency and Generation Time of Our Models}
  \label{tab:supp_latency}
  \begin{threeparttable}
    \begin{tabular}{lccc}
      \toprule
      Model & Param.(B) & \makecell{Per-step\\ Latency(ms)} & \makecell{4-step\\ Generation(s)} \\
      \midrule
      Ours-tiny  & 0.3 & 280 & 1.2 \\
      Ours-small & 0.4 & 360 & 1.8 \\
      Ours-full\tnote{*} & 1.6 & 1580 & 6.7 \\
      \bottomrule
    \end{tabular}
    “$*$”: 4-bit weight quantized
  \end{threeparttable}
\end{table}

\section{Demo on Mobile Device}
\begin{figure*}
\begin{center}
\includegraphics[width=0.9\linewidth]{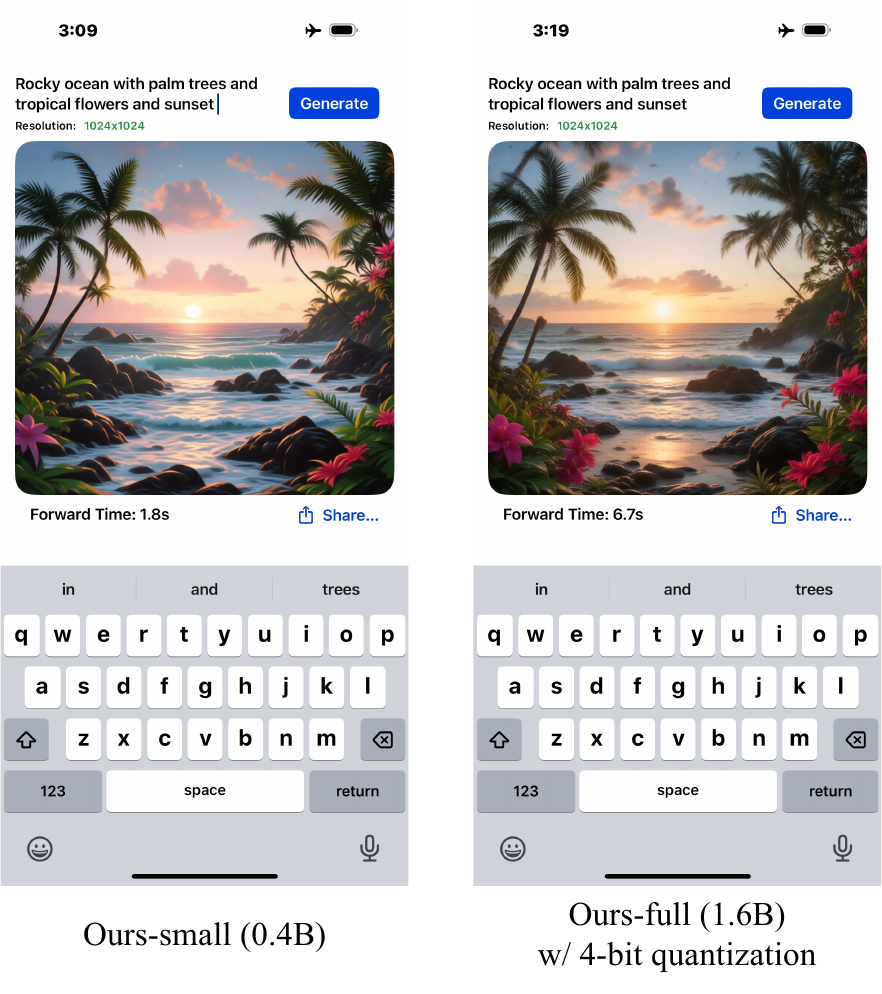}
\end{center}
\caption{\textbf{On-device Demo.}
  Screenshots from our on-device application running on an iPhone~16~Pro~Max. 
  The left panel shows results from the small model (0.4B), and the right panel shows results from the full variant (1.6B) with 4-bit quantization. }
\label{fig:supp_demo}
\end{figure*}

We include an \textbf{on-device demonstration} on the \href{https://snap-research.github.io/snapgenplusplus/}{project page}, showcasing our small model (0.4B). It achieves a generation time of $1.8\,s$ per image and produces high-quality outputs at 1024×1024 resolution on an iPhone~16~Pro~Max. The application is implemented using the open-source Swift Core ML Diffusers framework. Two screenshots of on-device generation on an iPhone~16~Pro~Max are shown in \cref{fig:supp_demo}, featuring results from our small and full variant with 4-bit quantization. Upon launching the app, users can input textual prompts and generate corresponding images by simply tapping the “Generate” button.

\newpage
\section{On-device Deployment Details}
To enable mobile-friendly deployment, we optimize the model to minimize computational overhead by reducing operations such as \texttt{transpose} and \texttt{reshape}. 
We structure the model in a convolutional fashion, where the channel dimension is placed as the third-to-last dimension (i.e., $(B, C, H, W)$), rather than following the conventional transformer layout of $(B, L, D)$.
We reimplement the attention mechanism using split \texttt{einsum} operations to improve on-device efficiency. 
For Blockwise Neighborhood Attention (BNA), computations for each block are executed in parallel through a for-loop, enabling efficient execution on mobile hardware. 
Finally, the model is exported via \texttt{CoreML} to generate a computation graph for deployment. \\

To deploy the full model (1.6B) on deivce, we quantize all linear and convolutional layer weights using k-means clustering over their values. Most layers are quantized to 4 bits (16 clusters), while more sensitive layers are assigned 8 bits. Sensitivity is determined with a simple heuristic: for each layer, we measure the mean-squared error (MSE) between the layer’s quantized output and the corresponding output from the unquantized model, when quantizing that layer in isolation. Layers with the largest degradation in MSE are designated as sensitive and quantized at 8 bits, resulting in an overall average quantization of 4.3 bits. After quantization, we freeze the weights and fine-tune the remaining parameters, such as biases and normalization layers, using self-distillation for several thousand iterations.

\section{Training Implementation Details}
We adopt FSDP2 for distributed training across 32 nodes, each equipped with 8 A100 GPUs (80\,GB). 
The model is initially trained at a resolution of $256^2$ with a global batch size of 8192 using the Adam optimizer and a learning rate of $1\times10^{-4}$ for 400K iterations under elastic training. 
Subsequently, the resolution is increased to $1024^2$ with a global batch size of 2048 and gradient checkpointing enabled. 
This stage incorporates knowledge distillation (KD) and continues under elastic training for an additional 100K iterations. \\

For the step-distillation stage (K-DMD), we set the time shift to 3, following the few-step teacher configuration in \cite{qwen_image_lightning_2025}. 
The teacher in the DMD objective employs $\text{cfg}=4$, consistent with the default setting of Qwen-Image~\cite{wu2025qwenimagetechnicalreport}. 
We apply LoRA to both the student network and the critic, using a rank of 64 and $\alpha=128$. 
The student is updated every 5 iterations. 
Training is conducted for 10K iterations across 4 nodes (global batch size 512) using the Adam optimizer with a learning rate of $1\times10^{-4}$ and $\beta=(0,\,0.99)$.

\section{Additional Illustration of Blockwise Neighborhood Attention }
\begin{figure*}
    \centering
    \includegraphics[width=\linewidth]{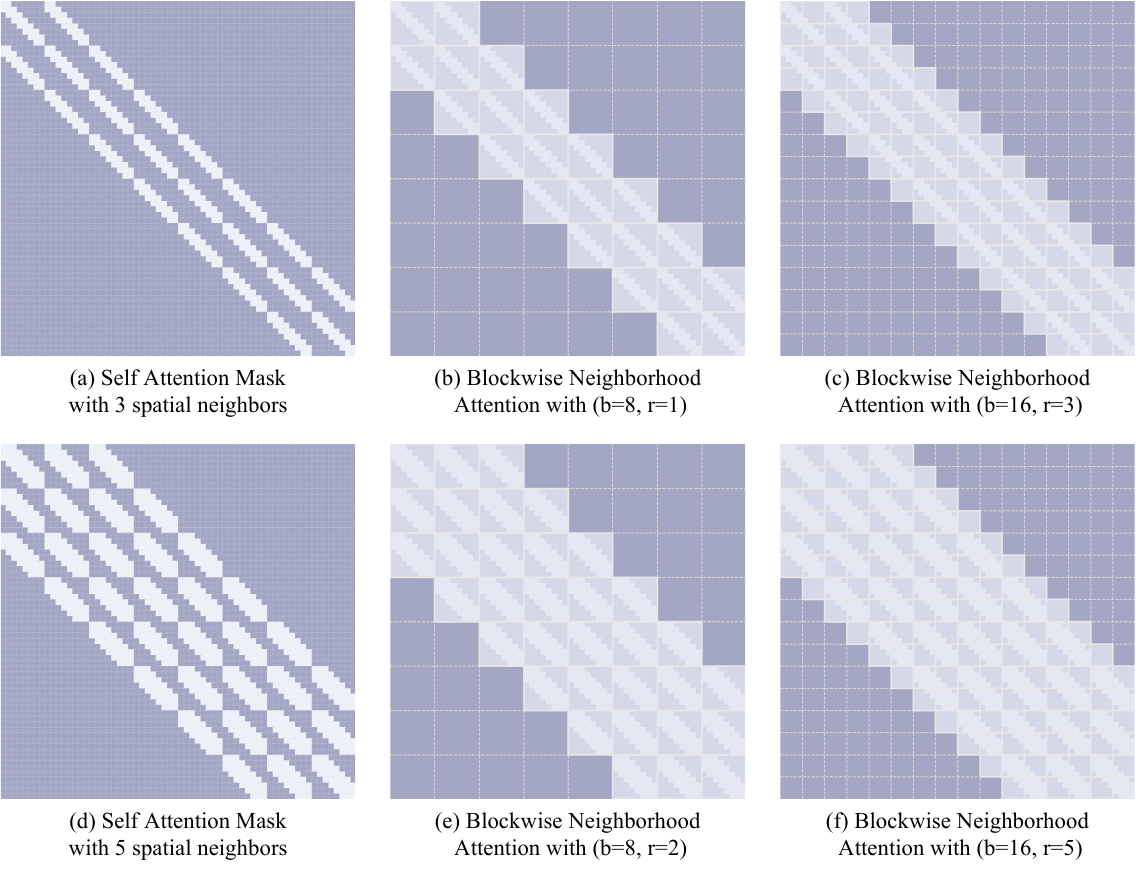}

\caption{\textbf{Illustration of Blockwise Neighborhood Attention.} 
Visualization of BNA under different hyperparameter settings of block number ($b$) and neighborhood radius ($r$), showing the corresponding spatial neighbor coverage and attention sparsity.}

\label{fig:supp_bna}
\end{figure*}

In \cref{fig:supp_bna}, we illustrate BNA under different configurations. Specifically, configurations (b) and (c) in BNA produce spatial neighbor coverage similar to the standard neighborhood attention with three token-level spatial neighbors in (a), while configurations (e) and (f) correspond closely to the five-neighbor case in (d). 
By adjusting the block number $b$ and neighborhood radius $r$, one can flexibly control the sparsity of BNA to balance computational efficiency and representational fidelity. In our experiments setting we set $b$ to 16 and $r$ to 1, essentially yields \textit{9} spatial neighbor tokens at $1024^2$ resolution.

\section{Detailed Results on T2I Benchmarks}
We present detailed results for DPG-Bench in 
\cref{tab:dpgbench}, GenEval in \cref{tab:geneval} and T2I-CompBench in \cref{tab:t2icompbench}.

\begin{table*}[!htbp]
\caption{\textbf{Detailed Results of DPG-Bench Comparisons.}}
\label{tab:dpgbench}
\centering
{\small
\begin{tabular}{lc|ccccc|c}
\toprule
Model & Param.(B) & Global & Entity & Attribute & Relation & Other & Overall $\uparrow$ \\
\midrule
SnapGen~\cite{hu2024snapgen}             & 0.4  & 88.3 & 85.1 & 87.0 & 87.3 & 87.6 & 81.1 \\ 
PixArt-$\alpha$~\cite{chen2023pixartalpha}  & 0.6  & 75.0 & 79.3 & 78.6 & 82.6 & 77.0 & 71.1 \\
PixArt-$\Sigma$~\cite{chen2024pixart}       & 0.6  & 86.9 & 82.9 & 88.9 & 86.6 & 87.7 & 80.5 \\
SANA~\cite{xie2025sana}                   & 1.6  & 86.0 & 91.5 & 88.9 & 91.9 & 90.7 & 84.8 \\
LUMINA-Next~\cite{zhuo2024lumina}         & 2.0  & 82.8 & 88.7 & 86.4 & 80.5 & 81.8 & 74.6 \\
SD3-Medium~\cite{esser2024sd3}            & 2.0  & 83.5 & 89.6 & 86.7 & 93.2 & 92.5 & 85.1 \\
SDXL~\cite{podell2023sdxl}                & 2.6  & 83.3 & 82.4 & 80.9 & 86.8 & 80.4 & 74.7 \\
Playgroundv2.5\cite{li2024playground}     & 2.6  & 83.1 & 82.6 & 81.2 & 84.1 & 83.5 & 75.5 \\
IF-XL~\cite{deepfloyd}                    & 5.5  & 77.7 & 81.2 & 83.3 & 81.8 & 82.9 & 75.6 \\
SD3.5-Large~\cite{sd3.5}                   & 8.1  & 87.4 & 92.1 & 90.0 & 88.2 & 88.1 & 85.6 \\
Flux.1-dev~\cite{blackforestlabs2024flux}  & 12   & 74.4 & 90.0 & 89.9 & 90.9 & 88.3 & 83.8 \\
HiDream-I1-Full~\cite{hidreami1technicalreport}& 17 & 76.4 & 90.2 & 89.5 & 93.7 & 91.8 & 85.9 \\
Qwen-Image~\cite{wu2025qwenimagetechnicalreport} & 20 & 91.3 & 91.6 & 92.0 & 94.3 & 92.7 & 88.3 \\
\midrule
Ours-tiny      & 0.3  & 88.5 & 90.2 & 88.8 & 92.6 & 78.8 & 84.6 \\ 
Ours-small     & 0.4  & 84.2 & 90.9 & 89.0 & 93.1 & 79.6 & 85.2 \\
Ours-full      & 1.6  & 85.7 & 91.5 & 89.6 & 94.5 & 80.4 & 87.2 \\ 
\bottomrule
\end{tabular}}
\end{table*}

\begin{table*}[!htbp]
\caption{ \textbf{Detailed Results of GenEval Bench Comparisons.}}
\label{tab:geneval}
\centering
\resizebox{1.0\linewidth}{!}{
\begin{threeparttable}[b]
\renewcommand{\arraystretch}{1.1}
{\small
\begin{tabular}{lc|cccccc|c}
\toprule
Model & Param.(B) &
\begin{tabular}[c]{@{}c@{}}Single \\ Object \end{tabular} &
\begin{tabular}[c]{@{}c@{}}Two \\ Objects \end{tabular} &
Counting &
Colors &
Position &
\begin{tabular}[c]{@{}c@{}}Color \\Attr. \end{tabular}  &
Overall $\uparrow$ \\ 
\midrule
SnapGen~\cite{hu2024snapgen} &  0.4   & 1.00   & 0.84   & 0.60  &  0.88    & 0.18   &  0.45 & 0.66 \\ 
PixArt-$\alpha$~\cite{chen2023pixartalpha}  & 0.6  & 0.98  & 0.50 & 0.44 & 0.80   & 0.08     & 0.07 & 0.48 \\
PixArt-$\Sigma$~\cite{chen2024pixart}       & 0.6  & 0.99  & 0.65 & 0.46 & 0.82   & 0.12     & 0.12 & 0.53 \\
SANA~\cite{xie2025sana}                    & 1.6  & 0.99  & 0.77   & 0.62 & 0.88  & 0.21  & 0.47 & 0.66 \\
LUMINA-Next~\cite{zhuo2024lumina}          & 2.0  & 0.92  & 0.46 & 0.48 & 0.70 & 0.09 & 0.13  & 0.46 \\
SD3-Medium~\cite{esser2024sd3}             & 2.0  & 0.98 & 0.74 & 0.63 & 0.67 & 0.34 & 0.36 & 0.62 \\
SDXL~\cite{podell2023sdxl}                 & 2.6  & 0.98  & 0.74 & 0.39     & 0.85   & 0.15     & 0.23 & 0.55 \\
Playgroundv2.5~\cite{li2024playground}     & 2.6  & 0.98  & 0.77 & 0.52     & 0.84   & 0.11     & 0.17 & 0.56 \\
IF-XL~\cite{deepfloyd}                    & 5.5  & 0.97  & 0.74   & 0.66 & 0.81  & 0.13  & 0.35  & 0.61 \\
SD3.5-Large~\cite{sd3.5}                   & 8.1  & 0.98 & 0.89 & 0.73 & 0.83 & 0.34 & 0.47 & 0.71 \\
FLUX.1-dev~\cite{blackforestlabs2024flux}  & 12   & 0.98 & 0.81 & 0.74 & 0.79 & 0.22 & 0.45 & 0.66 \\
HiDream-I1-Full~\cite{hidreami1technicalreport} & 17 & 1.00 & 0.98&0.79 &0.91 &0.60 &0.72 &0.83 \\
Qwen-Image~\cite{wu2025qwenimagetechnicalreport} & 20 & 0.99 &0.92 &0.89 &0.88 &0.76 &0.77 &0.87 \\
\midrule
Ours-tiny   & 0.3  & 1.00& 0.91 & 0.62 & 0.85 & 0.26  & 0.56 & 0.69 \\ 
Ours-small  & 0.4  & 1.00& 0.91 & 0.64 & 0.89 & 0.22  & 0.55 & 0.70 \\
Ours-full   & 1.6  & 1.00& 0.97 & 0.66 & 0.90 & 0.32 &0.70 & 0.76 \\ 
\bottomrule
\end{tabular}}
\end{threeparttable}
}
\end{table*}

\begin{table*}[!htbp]
\caption{\textbf{Detailed Results of T2I CompBench Comparisons.}}
\label{tab:t2icompbench}
\centering
\resizebox{1.0\textwidth}{!}{%
{\small
\begin{tabular}{lc|cccccc|c}
\toprule
Model & Param.(B) & Color & Complex & Nonspatial & Shape& Spatial & Texture & Overall$\uparrow$ \\
\midrule
PixArt-$\alpha$~\cite{chen2023pixartalpha}  & 0.6      & 0.416 & 0.334 & 0.308 & 0.389 & 0.197 & 0.461 & 0.351 \\
PixArt-$\Sigma$~\cite{chen2024pixart}       & 0.6     & 0.585 & 0.380 & 0.309 & 0.479 & 0.244 & 0.566 & 0.427 \\
SANA~\cite{xie2025sana}                    & 1.6       & 0.660 & 0.377 & 0.312 & 0.529 & 0.322 & 0.652 & 0.476 \\
LUMINA-Next~\cite{zhuo2024lumina}          & 2.0      & 0.511 & 0.350 & 0.303 & 0.333 & 0.185 & 0.438 & 0.353 \\
SD3-Medium~\cite{esser2024sd3}             & 2.0            & 0.794 & 0.384 & 0.315 & 0.582 & 0.324 & 0.731 & 0.522 \\
SDXL~\cite{podell2023sdxl}                 & 2.6            & 0.570 & 0.331 & 0.311 & 0.481 & 0.199 & 0.520 & 0.402 \\
Playgroundv2.5~\cite{li2024playground}     & 2.6  & 0.644 & 0.364 & 0.308 & 0.486 & 0.217 & 0.607 & 0.437 \\
IF-XL~\cite{deepfloyd}                    & 5.5           & 0.591 & 0.354 & 0.311 & 0.512 & 0.182 & 0.577 & 0.421 \\
SD3.5-Large~\cite{sd3.5}                   & 8.1           & 0.768 & 0.382 & 0.316 & 0.591 & 0.275 & 0.712 & 0.507 \\
FLUX.1-dev~\cite{blackforestlabs2024flux}  & 12           & 0.764 & 0.374 & 0.307 & 0.501 & 0.253 & 0.627 & 0.471 \\
HiDream-I1-Full~\cite{hidreami1technicalreport} & 17        & 0.749 & 0.401 & 0.314 & 0.592 & 0.399 & 0.696 & 0.525 \\
Qwen-Image~\cite{wu2025qwenimagetechnicalreport} & 20            & 0.836 & 0.399 & 0.317 & 0.605 & 0.443 & 0.743 & 0.557 \\
\midrule
Ours-tiny   & 0.3      & 0.765 & 0.372 & 0.316 & 0.545 & 0.331 & 0.680 & 0.502 \\
Ours-small  & 0.4      & 0.770 & 0.370 & 0.316 & 0.551 & 0.350 & 0.679 & 0.506 \\
Ours-full   & 1.6        & 0.794 & 0.375 & 0.316 & 0.600 & 0.419 & 0.712 & 0.536 \\
\bottomrule
\end{tabular}}
}
\end{table*}





\section{Qualitative Comparison on ImageNet}
\begin{figure*}[!htbp]
\begin{center}
\includegraphics[width=1\linewidth]{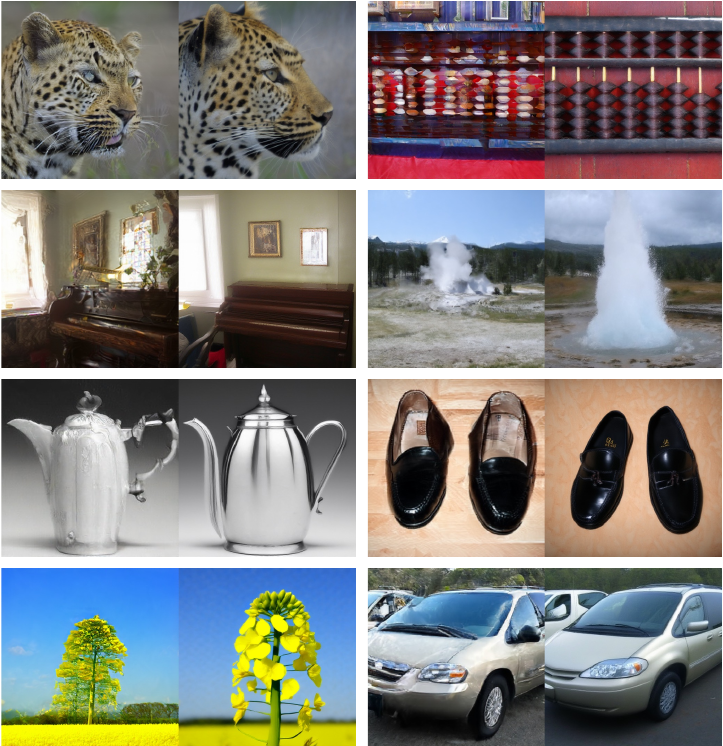}
\end{center}
\caption{\textbf{Qualitative comparison on ImageNet-1K.} 
Visual comparison between on-device models SnapGen~\cite{hu2024snapgen} (0.4B, left in each pair, Val Loss = 0.5131) and our small model (0.4B, right in each pair, Val Loss = 0.5090). 
Our model produces sharper textures, more consistent colors, and improved structural fidelity across diverse categories.}

\label{fig:supp_imagenet}
\end{figure*}

We present some visual results of ImageNet-1K between our 0.4B small model (Validation Loss = 0.5090) and SnapGen U-Net~\cite{hu2024snapgen} (0.4B, Validation Loss = 0.5131) in \cref{fig:supp_imagenet}. Our model produces sharper textures, more consistent colors, and improved structural fidelity across diverse categories.

\section{Additional Qualitative Comparison on T2I}
\begin{figure*}
    \centering
    \includegraphics[width=0.96\linewidth]{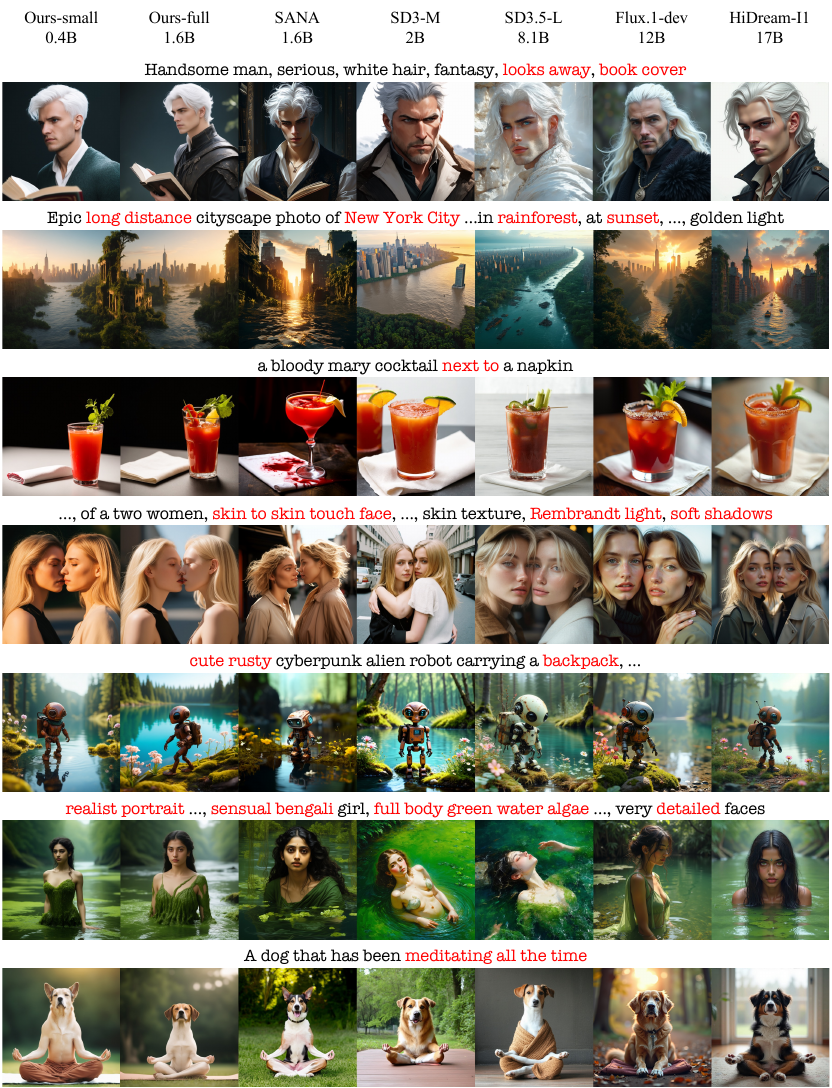}
     \caption{\textbf{Additional Qualitative Comparison.} Our models demonstrate competitive visual quality and superior prompt-following ability.  Input text prompts are shown above each image grid; all images are generated at $1024^2$ resolution. Zoom in for details.}
    \label{fig:supp_t2i_vis1}
\end{figure*}

\begin{figure*}
    \centering
    \includegraphics[width=0.96\linewidth]{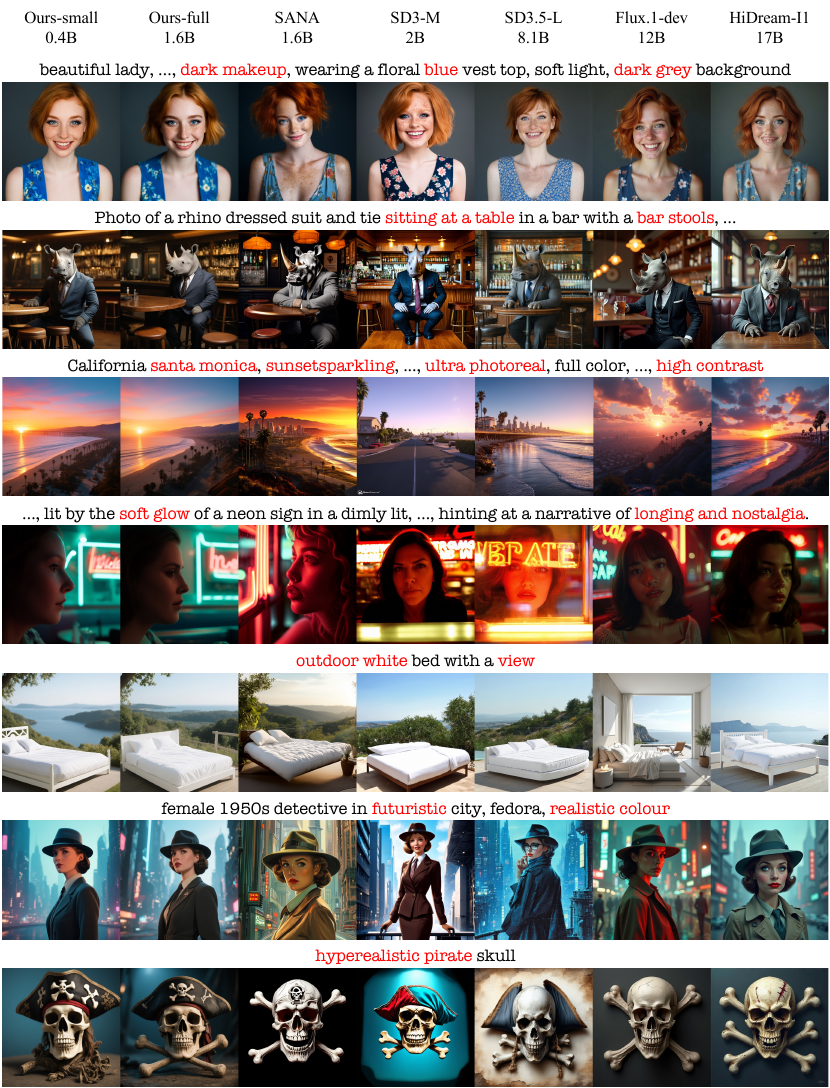}
    \caption{\textbf{Additional Qualitative Comparison.} Our models demonstrate competitive visual quality and superior prompt-following ability.  Input text prompts are shown above each image grid; all images are generated at $1024^2$ resolution. Zoom in for details.}
    \label{fig:supp_t2i_vis2}
\end{figure*}

To further demonstrate the visual fidelity and prompt adherence of our model, we provide additional qualitative comparisons on text-to-image (T2I) generation tasks.
Our models are evaluated across diverse prompts spanning objects, scenes, and artistic compositions, highlighting their ability to produce high-quality, semantically accurate, and visually consistent outputs.
As shown in \cref{fig:supp_t2i_vis1} and \cref{fig:supp_t2i_vis2}, our approach demonstrates competitive visual quality and strong alignment with textual descriptions, while achieving results comparable to or better than baseline methods with significantly larger parameter counts.

\end{document}